\pdfoutput=1

\documentclass[11pt]{article}

\usepackage[final]{acl}

\usepackage{times}
\usepackage{latexsym}
\usepackage{pgfplots}
\usepackage{lipsum}
\usepackage{changepage}
\usepackage{amsmath}
\usepackage[T1]{fontenc}

\usepackage[utf8]{inputenc}

\usepackage{microtype}

\usepackage{inconsolata}

\usepackage{graphicx}

\usepackage{subcaption}
\usepackage{multirow}
\usepackage{bbm}
\usepackage{bm}
\usepackage{enumitem}
\usepackage[most]{tcolorbox}
\usepackage{booktabs}
\usepackage[table]{xcolor}

\def\argmax{\operatornamewithlimits{arg\,max}}

\title{Mapping the Course for Prompt-based Structured Prediction}

\author{Matt Pauk \\
  University of Colorado Boulder \\
  \texttt{matt.pauk@colorado.edu} \\\And
  Maria Leonor Pacheco \\
  University of Colorado Boulder \\
  \texttt{maria.pacheco@colorado.edu} \\}

\begin{document}
\maketitle
\begin{abstract}
Large language models (LLMs) have demonstrated strong performance in a wide-range of language tasks without requiring task-specific fine-tuning. However, they remain prone to hallucinations and inconsistencies, and often struggle with complex reasoning, in part due to the limitations of autoregressive generation. We propose to address some of these issues, particularly for structured prediction, by combining LLMs with combinatorial inference to marry the predictive power of LLMs with the structural consistency provided by inference methods. We perform exhaustive experiments in an effort to understand which prompting strategies can best estimate confidence values for downstream symbolic inference, and find that, independent of prompting strategy, incorporating symbolic inference yields more consistent and accurate predictions than prompting alone. Finally, we show that calibration and fine-tuning with structured learning objectives further increases performance on challenging tasks, highlighting that structured learning remains valuable in the era of LLMs.

\end{abstract}

\section{Introduction}

Prompting large language models (LLMs) has been shown to be an effective methodology for a variety of natural language processing (NLP) tasks \citep{10.1162/coli_a_00523}. Through pre-training on massive-scale text corpora, general-purpose LLMs acquire extensive world knowledge that can be applied across tasks without specialized training \citep{fewshotlearners,wei2022chain}. This makes it possible to generate answers to a broad set of questions or instructions by conditioning the generation on a textual input prompt. 
As a result, prompting provides us with a flexible and adaptable framework for addressing new problems, where good performance can often be achieved simply by designing appropriate prompts and curating a handful of input-output demonstrations. However, regardless of the prompting strategy used, these models remain limited by their training objectives~\cite{Radford2018ImprovingLU,wei2022finetuned,NEURIPS2022_b1efde53}, which often lead to hallucinations and struggles with more complex reasoning tasks \citep{ji2023survey,doi:10.1073/pnas.2322420121}. One such area where these limitations become apparent is that of structured prediction. 


We use the term structured prediction to describe any machine learning task that consists of predicting several individual but related components as part of some structured object \citep{10.7551/mitpress/7443.001.0001}. These tasks are widely common in NLP, where we are often interested in parsing text into complex linguistic structures~\citep{mann_rhetorical_1988,Lascarides2007,palmer2010semantic,kamath2019a} or aligning language with structured knowledge~\citep{mao2018the,pacheco-goldwasser-2021-modeling}. Most previous work applying LLMs to structured prediction tasks treats each individual component of the structure as an independent prompting task \citep{roy-etal-2022-towards} or prompts the model to produce the entire structure all at once as a sequence of tokens~\citep{ettinger-etal-2023-expert}. However, because these strategies are solely relying on auto-regressive generation for predicting structures, they do not have a way to strictly enforce that the predicted structure is a valid one. 

Recently, \citet{mehta-etal-2024-promptly} proposed a framework for explicitly modeling dependencies by combining traditional inference algorithms with prompt-based predictions. Their approach first scores local candidate substructures by prompting LLMs and then finds the best global output via constrained optimization. They tested their framework on semantic role labeling and co-reference resolution, and showed that enforcing consistency improves performance over unconstrained prediction while guaranteeing structurally valid outputs. While these results show the relevance and potential of constrained inference for prompt-based structured prediction, it is not clear \textit{how} we should derive these scores from prompt-based inferences. In their work, \citet{mehta-etal-2024-promptly} simply take the raw likelihoods of generating particular responses and directly plug them into the maximum a posteriori (MAP) inference process. 

Traditional structured inference algorithms rely on weighting certainty scores for the different candidate substructures and finding the best global assignment that satisfies the structural constraints of the problem. Historically, trained ad hoc models generated label scores, which would correspond to the locally or globally normalized likelihood of the candidate label based on a learned conditional distribution $P(Y|X)$~\cite{10.5555/645530.655813,collins-2002-discriminative,Chang2012StructuredLW,pacheco-goldwasser-2021-modeling}. Unlike trained discriminative classifiers, prompt-based approaches do not learn a conditional distribution for the specific task, but rather estimate the probability of generating a token in a vocabulary given the sequence of tokens provided in the prompt~\cite{xie2022an}. This makes deriving certainty scores for prompt-based predictions very challenging. Given the wide range of approaches for certainty estimation~\cite{kadavath2022languagemodelsmostlyknow,xiong2024can} and calibration~\cite{jiang-etal-2021-know,tian-etal-2023-just} in LLMs, it is hard to determine which -if any- of them would be appropriate for structured inference algorithms. This is especially relevant given that many of these techniques have struggled to align confidence scores with actual accuracy and reliably predict failure cases~\cite{xiong2024can}.

In this paper, we are interested in systematically exploring and comparing different ways to score candidate substructures for prompt-based prediction, as well as exploring learning mechanisms to align LLMs with structured objectives. To this end, we present an exhaustive study of different ways to combine the predictive power of LLMs with structured inference to enforce the structural consistency of LLM predictions. Specifically, our goal is to answer the following research questions.

\begin{enumerate}[]
\item What is the most effective way to estimate confidence scores for individual output components from LLMs for use in combinatorial inference?
\item What is the best way to fine-tune LLMs for structured prediction tasks?
\end{enumerate}
\color{black}
We evaluated all explored methods on two challenging discourse-level structured prediction tasks: morality framing and coreference resolution, and show that inference helps performance, regardless of the strategy used to extract confidence scores. Of the confidence estimation strategies we explore, formulating the prompt as a true/false question performs the best. Additionally, we show that fine-tuning models using the global structured prediction objective leads to improved performance over fine-tuning on the individual local decisions. All the code needed to reproduce our experiments is available to the community\footnote{\url{https://github.com/mappauk/prompt-based-structured-inference}}. 


\section{Related Work}

We survey related work along three directions; deep structured prediction, structured prediction with LLMs, and confidence scoring and recalibration with LLMs.

\subsection{Deep structured prediction} There is a lot of previous work combining neural models with symbolic inference for structured prediction tasks. 
Neural networks globally normalized using structured inference have been successfully applied to sentence-level NLP tasks such as named entity recognition and dependency parsing~\cite{chen-manning-2014-fast, weiss-etal-2015-structured,ma-hovy-2016-end,lample-etal-2016-neural,kiperwasser-goldberg-2016-simple,malaviya-etal-2018-neural}. 

When dealing with tasks that go beyond sentence-level dependencies, most prior work combines the output scores of independently trained classifiers using inference~\cite{beltagy-etal-2014-probabilistic,ning-etal-2018-joint,pryor-etal-2023-using,leto-etal-2024-framing}
while others create ad hoc joint learning approaches for their particular tasks~\cite{han-etal-2019-joint,widmoser-etal-2021-randomized}. 

Our work most closely resembles \citet{pacheco-goldwasser-2021-modeling}, who propose a general framework for combining deep learning models with structured inference. However, rather than fine-tuning custom neural networks, we leverage general-purpose generative LLMs and few-shot prompting strategies.

\subsection{Structured prediction with LLMs} How best to use generative LLMs for structured prediction tasks is still a largely unexplored area. Most previous work either prompts the model to make predictions on local components without any enforcement of structural constraints \citep{roy-etal-2022-towards} or prompts the model to predict the entire structure at once \citep{ettinger-etal-2023-expert}. Along these lines, \citet{liu-etal-2022-autoregressive} propose an approach that models structures as sequences of actions in an autoregressive manner, without incurring any loss. 

The closest work to our proposed method is \citep{mehta-etal-2024-promptly}, who also combine few-shot prompting with combinatorial inference. However, they do not explore different LLM confidence estimation or calibration methods. Additionally, we explore learning strategies using global structured prediction objectives for prompt-based prediction, which to the best of our knowledge has not been explored before. 

\subsection{Confidence/uncertainty scoring and recalibration with LLMs} There is a considerable amount of research surrounding confidence estimation in LLMs. \citet{geng-etal-2024-survey} provide a survey of common confidence estimation strategies, considering both white-box methods that leverage internal model layers and/or token probabilities, as well as black-box methods that rely only on generated text. 

For white-box methods, \citet{kadavath2022languagemodelsmostlyknow} show that when prompts are formulated as true or false questions or multiple choice questions, the generated token probabilities themselves can be a well-calibrated form of confidence estimation. 

For black-box methods, \citet{manakul-etal-2023-selfcheckgpt} estimate confidence values by sampling a number of generations and use the consistency and variability between generations as a proxy of confidence. 

Alternatively, there is a large body of work on how to prompt and/or train models to estimate their own confidence level using numerical scales \citep{lin2022teaching,xiong2024can,liu-etal-2024-llms-learn-uncertainty}.
\section{Prompt-based Structured Prediction}~\label{sec:prompt-based-struct-prediction}
We outline a general framework for modeling any structured prediction task via prompting. To do this, we define each problem as an LLM-based factor graph $\Psi$ with potentials $\psi_i \in  \Psi$ over all possible structures $Y$. Each decision $\psi_i \in \Psi$ is scored using a decoder-only LLM instance $\rho$ with parameters $\bm\theta$. 

Let $\rho(\bm{x_i,y_i}; \bm\theta)$ be the score for the potential $\psi_i$ resulting from the prompting of an LLM $\rho$ parameterized by $\bm\theta$. In this way, we can find the optimal structure $\bm{y} \in Y$ by performing MAP inference as:

\begin{equation}
\begin{split}
\argmax_{\bm{y}\in Y} \sum_{\psi_i \in \Psi} \rho(\bm{x_i,y_i}; \bm\theta) \\
s.t. \; c(\bm{x_c,y_c}) \;\;\forall c \in C
\end{split}
\label{eq:inference}
\end{equation}

\noindent where $C$ is the set of structural constraints defined by the factor graph $\Psi$, and $\bm{x_c}, \bm{y_c}$ correspond to the inputs and variables relevant to the constraints. 

Following a long tradition on structured prediction for NLP, we use Integer Linear Programming (ILP) to calculate $\argmax$. Given that every MAP inference problem with discrete variables can be represented as a linear objective, ILP provides us with the utmost flexibility to represent any structured prediction problem~\cite{10.1145/1102351.1102444}. In addition, all Boolean functions can be compiled into a set of linear inequalities to be used as constraints in the ILP formulation~\cite{DBLP:journals/corr/abs-2307-00171}. Importantly, we note that this framework allows ILP to be replaced by any appropriate inference algorithm for a given task, including tractable linear programming relaxations such as $AD^3$~\cite{10346}. Our general-purpose ILP formulation is modeled after prior work \citep{pacheco-goldwasser-2021-modeling}, with specific implementation details outlined in Appendix \ref{appendix:inference}.
\section{Strategies for Scoring Sub-structures}~\label{sec:scoring-strategies}
The scores to be used with inference are obtained by prompting an LLM. However, it is unclear how best to extract probability values for classification tasks using LLM prompts~\citep{geng-etal-2024-survey}. Unlike a classifier explicitly trained to predict output probabilities, an LLM provides a probability distribution over all the tokens in it's vocabulary. Additionally, for many proprietary models, we do not even have access to the generated token probabilities, requiring a confidence estimation strategy that works using only the plain text generations. We experiment with an exhaustive list of confidence estimation methods proposed in the literature. Following \citet{geng-etal-2024-survey}, we divide these methods into white-box and black-box confidence estimation strategies.

\subsection{White-Box Methods}~\label{sec:white-bot-methods}
We experiment with several white-box confidence estimation methods, which we describe as requiring access to the token probability values for each prompt input and generation.

\subsubsection{True/False Token Prediction} Inspired by results showing that LLMs are fairly well calibrated for true/false and multiple choice questions \citep{kadavath2022languagemodelsmostlyknow}, we format the task as a true/false question and compute confidence estimation values based on the probability that the LLM generates the true token as $p(y=True|x)$.

In the multiclass case, the confidence value is computed by normalizing the probability of $y$ being true for a particular class $c$ over the sum of the true token probabilities for all classes as:
\begin{equation}
\frac{p(y=True|x_c)}{\sum_{i}^{|C|}{p(y=True|x_{c_i})}}
\end{equation}

\subsubsection{Multiple Choice} This strategy formulates the problem as a question-answering (QA) task with the possible labels given as multiple choice options. The score for a given text and label is then the probability of generating the token of the corresponding multiple choice option $o$ as
$p(y=o|x)$.

\subsubsection{Generative Classification}\label{sec:generative-classification} Instead of the standard QA format used by the True/False and Multiple Choice methods, where the model is asked to predict the label given some text to classify, we can flip the problem formulation on its head and provide the label as the input prompt and estimate the label likelihood based on the probability of the model to generate the example text. This approach has been shown to improve worst-case performance and reduce output variance~\cite{kumar2024genz}. Following prior work, we formulate the input prompt as \texttt{"\{natural language label description\} \{text to be classified\}"}. We then compute the score for a particular label using the following equation:
\begin{equation}\label{gz-equation}
\frac{1}{m}\sum_i^m \sum_j^n p(x_j| y_ix_1...x_{j - 1})
\end{equation}
Where $y_i$ represents one of $m$ versions of the label description and $x=x_1...x_n$ represents the text to be classified. Multiple versions of the language descriptions are used to reduce variance. The score calculated from Equation \ref{gz-equation} is then normalized over all possible labels to extract a final confidence value.

\subsection{Black-box Methods}~\label{sec:black-box-methods}
For many proprietary models, the token probabilities for the input prompt and / or the generations are not publicly available. For these models, the methods covered in Section \ref{sec:white-bot-methods} are not viable options. Therefore, we also explore several confidence estimation options that can be performed given only the plain-text generations of the LLM.

\subsubsection{Generation Sampling} This strategy uses consistency as a proxy for confidence by computing a score based on sampling generations, relying on the inherent randomness of these models. The model is prompted $n$ times to classify the text into one of the possible labels $L=l_1,l_2,....l_m$, giving us a set of $n$ generations $G=g_1,g_2,....g_n$. The number of generations that match a particular label $l_i$ is given by $G_{l_i} = \{g \in G | g_i = \mathtt{l_i}\}$. The score for label $l_i$ is then calculated as 
$\frac{|G_{l_i}|}{n}$.

\subsubsection{Verbalized Confidence} This method prompts the model to estimate its own confidence level in the answer. We use the prompting method proposed by \citet{xiong2024can}, where the model is given the text to be classified and one of the possible classes, and asked to estimate its confidence level in the answer on a scale of 0-100. We use the following prompt format: \\

\begin{adjustwidth}{1.5em}{1.5em} 
\small
\texttt{Question: \{q\}}\\
\texttt{Possible Answer: \{a\}}\\
\texttt{Q: How likely is the above answer to be correct? Do not elaborate on your answer or provide any explanation, answer only with the confidence value in the following format:}
\\
\texttt{Confidence: [the probability of answer {label} to be correct (0-100), not the one you think correct, please only include the numerical number in the range of 0-100]}
\end{adjustwidth}
\vspace{10pt}

This prompt is executed several times, and the final confidence value is given as the average confidence value elicited over all generations.
\section{Learning}~\label{sec:learning}
Sections \ref{sec:prompt-based-struct-prediction} and \ref{sec:scoring-strategies} describe a process to combine few-shot prompting strategies using pre-trained language models and combinatorial inference for structured prediction tasks. However, in the process described so far, there is no learning taking place, and we are solely relying on the world knowledge contained in the parametric space. In this section, we describe several fine-tuning strategies for the structured prediction task.

\subsection{Few-Shot Score Calibration} 

For this strategy, we train a logistic regression layer on top of the LLM scores to better calibrate them for the structured prediction task. The parameters of the LLM remain frozen and only the weights of the logistic regression layer are tuned. A separate logistic regression model $\phi_p$ is used for each prompting strategy $p$ within the structured prediction problem. We experiment with two mechanisms to train these regression models. 

The first, which we refer to as \textit{\textbf{Local Calibration}} involves tuning each logistic regression model separately for their respective sub-problem tasks using the cross-entropy loss function: 

\begin{equation}
L = -\log \sum_i^Cy_i\log(\hat{y_i})
\end{equation}

\noindent Where $\hat{y_i} = \phi_p(w)$ is the output of the logistic regression model for the strategy $p$, that takes as input the confidence scores $w$, extracted using one of the LLM prompting strategies in Section \ref{sec:scoring-strategies}. 

Alternatively, we jointly train all models $\phi_p$ using the structured hinge loss \citep{hume-structuredprediction}. To compute this loss, we perform structured inference as formulated in Section \ref{sec:prompt-based-struct-prediction}. However, instead of using the raw confidence scores $w$ from the LLM, we use the corresponding output of the logistic regression layer $s_i = \phi(w)_i$. The structured hinge loss can then be formulated as follows:

\begin{equation}~\label{structured-hinge-loss}
L = \max \left\{0, \sum_{\hat{y_i} \in \hat{y}} s_i\hat{y_i}   - \sum_{y_i \in y} s_iy_i \right\} 
\end{equation}

\noindent Where $\hat{y} \in Y$ is the current result of inference and $y \in Y$ corresponds to the gold structure. We refer to this mechanism as \textit{\textbf{Global Calibration}}.

\subsection{Local Fine Tuning} 

Supervised LLM fine-tuning has proven to be an effective method to improve LLM prompting performance for a specific task \citep{zhang2023instruction}. For this strategy, we fine-tune an LLM on the same prompts discussed in Section \ref{sec:scoring-strategies} using the standard loss for the next token prediction:

\begin{equation}
L = - \sum_{t=1}^{T}\log(p(y_t|x;y_1,y_2,....y_{t-1}))
\end{equation}

Fine-tuning is performed for each of the local components of the structured prediction problem. After fine-tuning the LLM, inference is performed by extracting confidence values using the same prompting strategy as used for fine-tuning. 

\begin{table*}[t]
\small
\vspace{0.2cm}
\centering
\begin{tabular}{llllllll}
 \hline
\textbf{Method} & \textbf{Shots} & \multicolumn{2}{c}{\textbf{Micro F1}}& \multicolumn{2}{c}{\textbf{Macro F1}}  & \multicolumn{2}{c}{\textbf{Constraint Violations}} \\
~ & ~ & \textbf{MF} & \textbf{Role} & \textbf{MF} & \textbf{Role} & \textbf{C1} & \textbf{C2}\\
 \hline
Few-shot ICL \citep{roy-etal-2022-towards} & 5 & 0.436 & n/a & n/a & n/a & n/a & n/a\\
 \hline
 \hline
 True/False & \multirow{2}{*}{5} & 0.498 & 0.444 & 0.466 & 0.378 & 1195 & 102 \\
 \hspace{3mm}+ constr & & \underline{0.517} & \textbf{\underline{0.452}} & \textbf{\underline{0.521}} & \textbf{\underline{0.408}} & 0 & 0 \\
 \hline
 Multiple Choice & \multirow{2}{*}{2} & 0.457 & 0.348 & 0.447 & 0.273 & 1133 & 234 \\
 \hspace{3mm}+ constr & & \textbf{\underline{0.530}} & \underline{0.386} & \underline{0.512} & \underline{0.321} & 0 & 0 \\
 \hline
 Generation Sampling & \multirow{2}{*}{5} & 0.389 & 0.387 & 0.419 & 0.276 & 1241 & 128 \\
 \hspace{3mm}+ constr & & \underline{0.456} & \underline{0.388} & \underline{0.447} & \underline{0.317} & 0 & 0 \\
 \hline
 Verbalized Confidence & \multirow{2}{*}{0} & 0.416 & 0.261 & 0.418 & 0.180 & 1275 & 312 \\
 \hspace{3mm}+ constr & & \underline{0.435} & \underline{0.322} & \underline{0.437} & \underline{0.274} & 0 & 0 \\
 \hline
 Generative Classification & \multirow{2}{*}{0} & 0.410 & 0.274 & 0.426 & 0.211 & 1072 & 338 \\
 \hspace{3mm}+ constr & & \underline{0.498} & \underline{0.295} & \underline{0.483} & \underline{0.245} & 0 & 0 \\
\hline
\end{tabular}
\caption{
Results for the Morality Frames task \citep{roy-etal-2021-identifying} for each of the five confidence elicitation methods (Section \ref{sec:scoring-strategies}), both with and without combinatorial inference. We present results only for best number of shots (0,2,5).}
\label{mf-main-results}
\end{table*}

\subsection{Global Fine-tuning} For this strategy, we use the same loss formulation as in the global calibration method described above. However, rather than freezing the LLM parameters, we backpropagate the structured hinge loss (Equation \ref{structured-hinge-loss}) into the LLM itself. Before global fine-tuning, we use local fine-tuning to hot-start the model parameters.
\begin{table*}[t]
\small
\vspace{0.2cm}
\centering
\resizebox{\textwidth}{!}{%
\begin{tabular}{lllll}
 \hline
\textbf{Model} & \multicolumn{2}{c}{\textbf{Micro F1}} & \multicolumn{2}{c}{\textbf{Macro F1}} \\
~ & \textbf{MF} & \textbf{Role} & \textbf{MF} & \textbf{Role}\\
\hline
\hline
GPT-J-6B Few Shot \citep{roy-etal-2022-towards} & n/a & n/a & 0.436 & n/a \\
Supervised Deep Structured Pred. \citep{roy-etal-2021-identifying} & n/a & n/a & 0.723 & 0.592 \\
Llama-8B Few Shot All In One & 0.415 & 0.374 & 0.456 & 0.357 \\
Llama-70B Few Shot All In One & 0.519 & 0.478 & 0.506 & 0.401 \\
GPT-5 Few Shot All in One & 0.616 & 0.625 & 0.577 & 0.535 \\
\hline
\hline
 Pre-trained few-shot baseline & 0.459 \textpm\ 0.01 & 0.345 \textpm\ 0.009 & 0.452 \textpm\ 0.023 & 0.281 \textpm\ 0.038 \\
 Pre-trained few-shot baseline (+ constr) & 0.529 \textpm\ 0.007 & 0.388 \textpm\ 0.014 & 0.511 \textpm\ 0.013 & 0.318 \textpm\ 0.015 \\
 
 Pre-trained Local Calibration & 0.522 \textpm\ 0.017 & 0.478 \textpm\ 0.006 & 0.450 \textpm\ 0.031 & 0.360 \textpm\ 0.040\\
 Pre-trained Local Calibration (+ constr) & 0.649 \textpm\ 0.006 & 0.562 \textpm\ 0.023 & 0.589 \textpm\ 0.019 & 0.453 \textpm\ 0.035 \\
 Pre-trained Global Calibration & \underline{0.660} \textpm\ 0.024 & \underline{0.594} \textpm\ 0.031 & \underline{0.638} \textpm\ 0.034  & \underline{0.512} \textpm\ 0.052\\
 
 \hline
 Local Fine-tuned baseline & 0.710 \textpm\ 0.013 & 0.691 \textpm\ 0.024 & 0.681 \textpm\ 0.018 & 0.595 \textpm\ 0.021 \\
 Local Fine-tuned baseline (+ constr) & 0.754 \textpm\ 0.005 & 0.731 \textpm\ 0.013 & 0.720 \textpm\ 0.008 & 0.656 \textpm\ 0.015\\
 Local Fine-tuned + Local Calibration & 0.724 \textpm\ 0.019 & 0.700 \textpm\ 0.025 & 0.695 \textpm\ 0.032 & 0.610 \textpm\ 0.020 \\
 Local Fine-tuned + Local Calibration (+ constr) & 0.759 \textpm\ 0.009 & 0.727 \textpm\ 0.013 & 0.722 \textpm\ 0.028 & 0.652 \textpm\ 0.019 \\
 Local Fine-tuned + Global Calibration & 0.758 \textpm\ 0.005 & 0.725 \textpm\ 0.008 & 0.722 \textpm\ 0.020 & 0.658 \textpm\ 0.026 \\
 Global Fine Tuning & \textbf{\underline{0.764}} \textpm\ 0.009 & \textbf{\underline{0.732}} \textpm\ 0.011 & \textbf{\underline{0.731}} \textpm\ 0.023 & \textbf{\underline{0.662}} \textpm\ 0.018 \\
\hline
\end{tabular}}
\caption{
Results on the Morality Frames dataset \citep{roy-etal-2021-identifying} after fine-tuning the LLMs. The multiple choice strategy for confidence estimation is used for all methods. 
}
\label{mf-learning-results}
\end{table*}

\section{Evaluation}
We evaluated our framework on two complex discourse-level tasks; morality framing and coreference resolution. For coreference, we use the GENIA Coreference biomedical dataset \citep{su2008coreference} and the CoNLL 2012 OntoNotes dataset \citep{pradhan2012conll}. For morality framing, we use the dataset released by~\citet{roy-etal-2021-identifying}. All datasets considered consist exclusively of English language text.

\subsection{Morality Framing in Political Tweets}

This task focuses on identifying the moral attitudes that are expressed in tweets made by members of the United States Congress~\citep{roy-etal-2021-identifying}. There are two aspects to the task; the first is identifying which of the five moral foundations (Care/Harm, Fairness/Cheating, Loyalty/Betrayal, Authority/Subversion, and Purity/Degradation) are being expressed in the tweet~\cite{fd3cbcce-5e3f-39a6-b275-9a1d627a0c23,Haidt2007-HAIWMO}. The other aspect of the task is to identify the moral role that the entities mentioned in the tweet are playing. As an example, consider the following tweet that expresses the moral foundation \textit{\textbf{Care/Harm}}: \\

\begin{adjustwidth}{1.5em}{1.5em}
\textit{This common-sense bill will reduce unnecessary and duplicative burdens on health care providers and patients in need of home health services}
\end{adjustwidth} 
\vspace{10pt}

The entity "common-sense bill" expresses the role of \textit{\textbf{entity providing care}}, "health care providers and patients" express the role of \textit{\textbf{target of care/harm}} and "duplicative burdens" express the role of \textit{\textbf{entity causing harm}}. We define prompt templates for both the moral role classification and moral foundation identification subproblems. Specific details on the prompt templates used can be found in Appendix \ref{appendix:mf-prompt-details}. After prompting, we perform inference to find the best global label assignments across both subproblems, subject to: \\


\begin{adjustwidth}{1.5em}{1.5em}
\noindent \textit{Constraint 1:} The predicted role of an entity in a tweet must align with the moral foundation predicted for the tweet.
\end{adjustwidth}
\begin{adjustwidth}{1.5em}{1.5em}
\noindent \textit{Constraint 2:} No two entities within the same tweet can be assigned the same role.
\end{adjustwidth}

\vspace{10pt}

\paragraph{Experimental Settings} We experiment with Llama-3.1-8B-Instruct \citep{dubey2024llama} and Mistral-7B-Instruct-v0.2 \citep{jiang2023mistral7b} as our base models. For fine-tuning experiments, we use LoRA \citep{hu2022lora} in combination with the Llama-3.1-8B-Instruct model. Details of our hyperparameter selection and hardware used can be found in Appendices \ref{appendix:mf-hyperparam} and \ref{app:hardware}, respectively. We evaluated performance by reporting macro and micro F1 scores for both moral foundation and role prediction. All results are averaged over five folds. 

\begin{table}[t]
\centering
\resizebox{\columnwidth}{!}{%
\begin{tabular}{lllll}
\hline
\textbf{Model} & \multicolumn{2}{c}{\textbf{Macro F1}} & \multicolumn{2}{c}{\textbf{Violations}} \\
~ & \textbf{MF} & \textbf{Role} & \textbf{C1} & \textbf{C2}\\
\hline
\hline
True/False - Llama-8B & 0.466 & 0.378 & 1195 & 102 \\
\hspace{3mm} + constr & \underline{0.517} & \underline{0.408} & 0 & 0 \\
\hline
True/False - Llama-70B & 0.446  & 0.318 & 1113 & 50 \\
\hspace{3mm} + constr & 0.491 & 0.404 & 0 & 0 \\
\hline
\hline
Multiple Choice - Llama-8B & 0.447 & 0.273 & 1133 & 234 \\
\hspace{3mm} + constr & \underline{0.512} & \underline{0.321} & 0 & 0 \\
\hline
Multiple Choice - Llama-70B & 0.407  & 0.288 & 767 & 124 \\
\hspace{3mm} + constr & 0.423 & 0.323 & 0 & 0 \\
\hline
\hline
Generation Sampling - Llama-8B & 0.419 & 0.276 & 1241 & 128 \\
\hspace{3mm} + constr & 0.447 & 0.317 & 0 & 0 \\
\hline
Generation Sampling - Llama-70B & 0.454  & 0.361 & 1092 & 83 \\
\hspace{3mm} + constr & 0.494 & 0.399 & 0 & 0 \\
\hline
Generation Sampling - GPT-5 & 0.563 & 0.456 & 735 & 82 \\
\hspace{3mm} + constr & \textbf{\underline{0.567}} & \textbf{\underline{0.473}} & 0 & 0 \\
\hline
\hline
\end{tabular}}
\caption{Results on the Morality Frames dataset \citep{roy-etal-2021-identifying} with varying model sizes.}\label{mf-large-experiments}
\end{table}

\begin{figure*}
\centering
\begin{subfigure}[b]{0.57\textwidth}
\resizebox{\columnwidth}{!}{%
\begin{tikzpicture}
    \begin{axis}[
        xlabel=Percentage Training Data,
        ylabel=Foundation Macro F1,
        legend style={at={(1.5,0.42)},anchor=south},
        xmin=1, xmax=5,
        ymin=0, ymax=1,
        ytick={0, 0.2, 0.4, 0.6, 0. 8, 1},
        xtick={1, 2, 3, 4, 5},
        xticklabels={1, 5, 20, 50, 100},
        grid=both,
        grid style={dashed,gray},
        minor grid style={dashed, gray},
        major grid style={dashed, gray}
        ]
    \addplot[smooth,mark=*,blue] plot coordinates {
        (1,0.207)
        (2,0.339)
        (3,0.430)
        (4,0.434)
        (5,0.435)
    };
    \addlegendentry{Local Calib}
    \addplot[dashed,mark=*,blue] plot coordinates {
        (1,0.210)
        (2,0.384)
        (3,0.541)
        (4,0.549)
        (5,0.596)
    };
    \addlegendentry{Local Calib (+ constr)}
    \addplot[smooth,mark=*,green] plot coordinates {
        (1,0.224)
        (2,0.416)
        (3,0.560)
        (4,0.602)
        (5,0.620)
    };
    \addlegendentry{Global Calib}
    \addplot[smooth,mark=*,red] plot coordinates {
        (1,0.491)
        (2,0.486)
        (3,0.656)
        (4,0.685)
        (5,0.683)
    };
    \addlegendentry{Local Fine Tuning}
    \addplot[dashed,mark=*,red] plot coordinates {
        (1,0.543)
        (2,0.496)
        (3,0.704)
        (4,0.710)
        (5,0.722)
    };
    \addlegendentry{Local Fine Tuning (+ constr)}
    \addplot[smooth,mark=*,black] plot coordinates {
        (1,0.542)
        (2,0.531)
        (3,0.704)
        (4,0.690)
        (5,0.744)
    };
    \addlegendentry{Global Fine Tuning}
    \end{axis}
\end{tikzpicture}}
\end{subfigure}
\begin{subfigure}[b]{0.35\textwidth}
\resizebox{\columnwidth}{!}{%
\begin{tikzpicture}
    \begin{axis}[
        xlabel=Percentage Training Data,
        ylabel=Role Macro F1,
        legend style={at={(1.5,0.44)},anchor=south},
        xmin=1, xmax=5,
        ymin=0, ymax=1,
        xtick={1, 2, 3, 4, 5},
        xticklabels={1, 5, 20, 50, 100},
        grid=both,
        grid style={dashed,gray},
        minor grid style={dashed, gray},
        major grid style={dashed, gray}
        ]
    \addplot[smooth,mark=*,blue] plot coordinates {
        (1,0.085)
        (2,0.233)
        (3,0.299)
        (4,0.326)
        (5,0.338)
    };
    \addplot[dashed,mark=*,blue] plot coordinates {
        (1,0.103)
        (2,0.270)
        (3,0.410)
        (4,0.405)
        (5,0.462)
    };
    \addplot[smooth,mark=*,green] plot coordinates {
        (1,0.086)
        (2,0.300)
        (3,0.428)
        (4,0.450)
        (5,0.480)
    };
    \addplot[smooth,mark=*,red] plot coordinates {
        (1,0.248)
        (2,0.211)
        (3,0.455)
        (4,0.518)
        (5,0.600)
    };
    \addplot[dashed,mark=*,red] plot coordinates {
        (1,0.333)
        (2,0.237)
        (3,0.571)
        (4,0.580)
        (5,0.634)
    };
    \addplot[smooth,mark=*,black] plot coordinates {
        (1,0.325)
        (2,0.304)
        (3,0.563)
        (4,0.563)
        (5,0.641)
    };
    \end{axis}
\end{tikzpicture}}
\end{subfigure}
\caption{Morality Framing results for each training strategy over varying amounts of training data}
\label{mf-varying}
\end{figure*}

\paragraph{Results} The results of the few-shot prompting methods are shown in Table \ref{mf-main-results}. We experiment with 0, 2, and 5 shots for each method except for the generative classification and verbalized confidence methods, which are zero-shot methods. Table \ref{mf-main-results} only contains the best result for each method using the Llama model, as it outperformed the mistral model in all strategies. The full results for both models can be found in Appendix \ref{appendix:extended-mf-results}. Overall, we see that inference (+ constr) improves performance compared to few-shot prompting alone, regardless of the confidence estimation strategy used. The best results are achieved when using the true/false method for extracting confidence scores; this holds both before and after inference. However, the multiple choice method is comparable to true/false for the moral foundation subproblem. Furthermore, our results outperform previous prompting work on the same task \citep{roy-etal-2022-towards}.

Table \ref{mf-learning-results} shows the results of our different calibration and fine-tuning strategies (Section \ref{sec:learning}). We use the multiple choice confidence estimation strategy for all fine-tuned models. Despite slightly worse performance than the True/False method, it requires much fewer prompts per instance, allowing for larger batch sizes during training, which are required when fine-tuning using the global structured objective (Equation \ref{structured-hinge-loss}). Similarly to the few-shot results, we find that adding inference helps performance regardless of the learning strategy. We notice that calibrating scores using logistic regression performs best when using Global Calibration. This finding holds true regardless of whether we use a pre-trained or fine-tuned LLM. However, the advantage of global calibration is much clearer in the pre-trained case. Unsurprisingly, local fine-tuning of the LLM leads to large performance gains over the few-shot version of the model.  

The best performing model uses the global fine-tuning method on top of the locally fine-tuned model, outperforming the previous state-of-the-art in this task \citep{roy-etal-2021-identifying}, which uses a classical deep structured prediction approach. We also compare our systems to different LLMs of varying size that attempt to predict the entire structure all at once with verbalized versions of our constraints in the prompt (Few Shot All In One). We find that the all-in-one prediction is an effective method that, at least in the case of Llama-70B and GPT-5, outperforms our baseline 8B models that use structured inference to enforce constraints. However, these all-in-one methods are outperformed by global calibration methods and are significantly outperformed by global fine tuning.

Table \ref{mf-large-experiments} displays the results of varying model sizes used to obtain priors for inference. We show results for the two best performing white-box methods (True/False and Multiple Choice) and the best performing black-box method (Generation Sampling). For each method, we show results for the Llama 8B and 70B models, and for the generation sampling method, we also experiment with the usage of GPT-5. We see that regardless of model size, the use of structured inference on top of the local predictions results in improved performance. Somewhat surprisingly, we notice that the 70B version of the Llama model actually performs slightly worse than the 8B model for the Multiple Choice and True/False methods. We suspect that the reason for this is that the additional parameters are likely encoding the ability to follow instructions for more complex tasks and, therefore, do not lead to increased performance on focused true/false or multiple-choice question-answering tasks.

Finally, Figure \ref{mf-varying} shows the effects of varying amounts of training data. For the most part, the same strategies emerge as the top performers in both the low-data and high-data regimes. We notice that with very low amounts of training data (1\%), all methods that fine-tune the parameter space have a larger advantage. Finally, we note that when imposing constraints (either during learning or prediction), the models converge with as little as 20\% of the training data.

\paragraph{Error Analysis} We find that some strategies are more biased than others towards certain moral foundations (see confusion matrices in Appendix~\ref{appendix:error-analysis}). In particular, verbalized confidence and generation sampling show a particular bias towards the authority/subversion class. 
In addition, we are able to find concrete evidence of structured inference correcting wrong local predictions. Consider the example shown in Figure~\ref{fig:example}, where we were able to align the predictions to be consistent with each other. That is, making sure that the moral roles assigned to entities are consistent with the overall moral foundation and that each entity plays a distinct role.

\begin{figure}[t]
\begin{tcbraster}[raster columns=1,raster equal height=rows,raster valign=top, size=small]
\begin{tcolorbox}[colback=white,colframe=black, title={Authority/Subversion}]
\small{
Rep. Greg Walden meets with C.O. veterans in Bend to discuss issues with \textcolor{blue}{[health care program{$]_{providing\_care}$}} to help \textcolor{purple}{[rural vets{$]_{being\_loyal}$}}}
\end{tcolorbox}  
\begin{tcolorbox}[colback=white,colframe=black, title={Care/Harm}]
\small{
Rep. Greg Walden meets with C.O. veterans in Bend to discuss issues with \textcolor{blue}{[health care program{$]_{providing\_care}$}} to help \textcolor{blue}{[rural vets{$]_{target\_care}$}}}
\end{tcolorbox}  
\end{tcbraster}
\caption{Predictions before (top) and after (bottom) structured inference.}\label{fig:example}
\end{figure}

\subsection{Coreference Resolution}

Co-reference resolution is the task of identifying whether or not two mentions of an entity within a piece of text refer to the same entity . We break the task down into a series of subproblems, where we prompt the LLM to determine whether two entity mentions within a document are coreferent or not. Prompt template details can be found in Appendix \ref{appendix:coref-prompt-details}. We define one custom hard constraint enforcing the transitivity of coreferent entity pairs as: \\

\begin{adjustwidth}{1.5em}{1.5em}
\noindent \textit{Constraint 1:} If entities A and B are coreferent and entities B and C are coreferent, then entities A and C must be coreferent.
\end{adjustwidth}

\vspace{10pt}

\paragraph{Experimental Settings} We experiment with Llama-3.1-8B-Instruct \citep{dubey2024llama} and Mistral-7B-Instruct-v0.2 \citep{jiang2023mistral7b} as our base models. For fine-tuning experiments, we use LoRA \citep{hu2022lora} in combination with the Mistral-7B-Instruct-v0.2 model. Details of our hyperparameter selection and hardware use can be found in Appendices \ref{appendix:coref-hyperparam} and \ref{app:hardware}, respectively. We evaluated performance by reporting macro F1 scores.

\begin{table}[t]
\centering
\resizebox{\columnwidth}{!}{%
\begin{tabular}{llll}
\hline
\textbf{Method} & \textbf{Shots} & \textbf{F1} & \textbf{Viol.} \\
\hline
\hline
Macaw-3B (+ constr) \citet{mehta-etal-2024-promptly} & 0 & 0.522 & 0 \\
\hline
\hline
 True/False - Mistral & \multirow{2}{*}{5} & 0.815 & 15212 \\
 \hspace{3mm}+ constr & & \underline{0.820} & 0 \\
 \hline
 Multiple Choice - Llama & \multirow{2}{*}{5} & 0.801 & 24928 \\
 \hspace{3mm}+ constr & & \underline{0.830} & 0 \\
 \hline
 Generation Sampling - Llama & \multirow{2}{*}{5} & 0.834 & 15086 \\
 \hspace{3mm}+ constr & & \textbf{\underline{0.842}} & 0 \\
 \hline
 Verbalized Confidence - Mistral & \multirow{2}{*}{0} & 0.506 & 25872 \\
 \hspace{3mm}+ constr & & \underline{0.512} & 0 \\
 \hline
 Generative Classification - Mistral & \multirow{2}{*}{0} & \underline{0.397} & 47794 \\
 \hspace{3mm}+ constr & & 0.371 & 0 \\
 \hline
 \hline
\end{tabular}}
\caption{
Co-reference results for the OntoNotes dataset \citep{pradhan2012conll} across all five prompting strategies}\label{coref-results-ontonotes}
\end{table}

\begin{table}[t]
\centering
\resizebox{\columnwidth}{!}{%
\begin{tabular}{llll}
\hline
\textbf{Method} & \textbf{Shots} & \textbf{F1} & \textbf{Viol.} \\
\hline
\hline
Flan-T5 (+ constr) \citet{mehta-etal-2024-promptly} & 0 & 0.654 & 0 \\
\hline
\hline
 True/False - Mistral & \multirow{2}{*}{5} & 0.799 & 29808 \\
 \hspace{3mm}+ constr & & \textbf{\underline{0.823}} & 0 \\
 \hline
 Multiple Choice - Mistral & \multirow{2}{*}{5} & 0.721 & 72303 \\
 \hspace{3mm}+ constr & & \underline{0.759} & 0 \\
 \hline
 Generation Sampling - Mistral & \multirow{2}{*}{5} & 0.747 & 48570 \\
 \hspace{3mm}+ constr & & \underline{0.781} & 0 \\
 \hline
 Verbalized Confidence - Mistral & \multirow{2}{*}{0} & \underline{0.599} & 123272 \\
 \hspace{3mm}+ constr & & 0.577 & 0 \\
 \hline
 Generative Classification - Llama & \multirow{2}{*}{0} & \underline{0.357} & 364712 \\
 \hspace{3mm}+ constr & & 0.305 & 0 \\
 \hline
 \hline
\end{tabular}}
\caption{
Co-reference results for the GENIA dataset \citep{su2008coreference} across all five prompting strategies.
}\label{coref-results-genia}
\end{table}

\begin{table}[t]
\centering
\resizebox{\columnwidth}{!}{%
\begin{tabular}{ll}
\hline
\textbf{Method} & \textbf{Macro F1}\\
\hline
\hline
 Fine-tuned Constrained Macaw-3B \citet{mehta-etal-2024-promptly} & 0.916 \\
\hline
\hline
 Pre-trained few-shot baseline & 0.721\\
 Pre-trained few-shot baseline (+ constr) & 0.759 \\
 Pre-trained Local Calibration & 0.771\\
 Pre-trained Local Calibration (+ constr) & 0.800\\
 Pre-trained Global Calibration & \underline{0.805}\\
 \hline
 Local Fine-tuned baseline & 0.883 \\
 Local Fine-tuned baseline (+ constr) & 0.887 \\
 Local Fine-tuned + Local Calibration & 0.882 \\
 Local Fine-tuned + Local Calibration (+ constr) & \textbf{\underline{0.891}} \\
 Local Fine-tuned + Global Calibration & 0.881 \\
 Global Fine Tuning & 0.890\\
 \hline
 \hline
\end{tabular}}
\caption{
Fine-tuning results on the GENIA coreference task \citep{su2008coreference}. The multiple choice confidence elicitation strategy is used for all models. 
}\label{coref-results-learning}
\end{table}

\paragraph{Results} Few-shot results for OntoNotes and GENIA are shown in Tables \ref{coref-results-ontonotes} and \ref{coref-results-genia}, respectively. We experiment with both Llama and Mistral instruct models and use 0 and 5 shots for all methods except for the zero-shot methods (verbalized confidence and generative classification). We report only the best model and the best number of shots for each strategy. The full results can be found in Appendix \ref{appendix:extended-coref-results}. We see that for all methods, except for the zero-shot ones, inference improves on the baseline results. We suspect that this is because the baseline model performance is so poor in these cases that the application of the transitivity constraint is ineffective. We find that the True/False, Multiple Choice, and Generation Sampling strategies have comparable performance for both datasets. 

We test our fine-tuning strategies on the GENIA dataset and again leverage the multiple choice strategy for its balance of performance and efficiency. The results are shown in Table \ref{coref-results-learning}. Similarly to the morality framing task, we note that structured inference improves performance in all scenarios. However, in the fine-tuned case, it is less helpful as performance on the task is already very high. 

\subsection{Computational Cost Experiments}
Tables \ref{comp-cost-mf} and \ref{comp-cost-genia} show the results of experiments measuring the computational efficiency of different methods when predicting a single structure for each task. We note that regardless of task, the Verbalized Confidence and Generative Classification methods are by far the most computationally expensive methods as they require multiple prompts for every candidate of a particular sub-task. This effect is even greater in the morality frames task when compared to coreference, as coreference only has two possible classes for a single subproblem, while morality frames has 16 possible moral roles for each entity and five possible moral foundations. Multiple choice is the least expensive method as it requires only a single prompt for each subproblem and is not affected by the number of classes associated with a particular subproblem. We also note that the cost of symbolic inference on top of prompting is negligible, even in the case of the most efficient prompting strategy (multiple choice).

As for the computational cost of training, the best supervised fine tuned coreference model took 13 hours and 37 minutes to train, and the best supervised fine tuned morality frames model took 5 hours and 7 minutes to train all five folds.

\begin{table}[t]
\centering
\begin{tabular}{ll}
\hline
\textbf{Method} & \textbf{Prompt Time (s)} \\
\hline
 True/False & 12.15\\
 Multiple Choice & 1.72 \\
 Generation Sampling& 12.03 \\
 Verbalized Confidence & 179.42 \\
 Generative Classification & 274.44 \\
 \hline
\end{tabular}
\caption{
Computational cost experiment results on the Morality Frames task \citep{roy-etal-2021-identifying}. We measure the amount of time it takes to predict the moral frame of a tweet and the moral roles of three entities mentioned in the tweet. Regardless of strategy we find that inference adds on average 0.11 seconds.
}\label{comp-cost-mf}
\end{table}

\begin{table}[t]
\centering
\begin{tabular}{ll}
\hline
\textbf{Method} & \textbf{Prompt Time (s)} \\
\hline
 True/False & 25.11 \\
 Multiple Choice & 12.42 \\
 Generation Sampling & 120.96 \\
 Verbalized Confidence & 535.20 \\
 Generative Classification & 321.14 \\
 \hline
\end{tabular}
\caption{
Computational cost experiment results on the GENIA coreference task \citep{su2008coreference}. We measure the amount of time it takes to predict coreference for all pairs of entities within a document using each strategy (154). Regardless of strategy we find that inference adds on average 3.03 seconds.
}\label{comp-cost-genia}
\end{table}

\color{black}
\section{Conclusion}

We show that structured inference is a useful tool for structured prediction tasks with LLMs, leading to consistent gains in performance over prompting alone. This holds true regardless of what strategy we use to extract confidence values from LLMs, although structuring the prompt as a true or false question works the best. Furthermore, we show that calibrating LLMs based on global structured prediction objectives can further boost performance. This finding holds true both for pre-trained models and models fine-tuned on the task. In the future, we want to explore porting the lessons learned here into multi-agent workflows, where inter-dependencies are observed, but sub-tasks are open-ended and supervision is not readily available.

\section*{Limitations}
We are limited on the size of the model that we can use for our fine-tuning experimentation. We report results on 7B and 8B parameter models for Mistral and Llama, but are unable to experiment with 70B+ parameter versions of these models when fine-tuning due to computational resource limitations. However, we experiment with larger parameter models for our pretrained, few-shot experiments (Llama-70B and GPT-5).

Lastly, we chose two representative tasks; the Morality Frames task, which provides a difficult relational reasoning task, and coreference resolution, which is a more traditional NLP task. While we believe that the evidence provided is sufficient to support our claims, future work could expand this task selection to provide greater evidence of generalization of these methods.
\section*{Ethical Considerations}

To the best of our knowledge, we did not violate the ACL code of ethics during the course of our work. We use existing, public datasets for evaluation and report model hyperparamaters and hardware details, as well as details of our prompting strategies to allow for the reproduction of our experiments. In addition, the code used to run our experiments has been made publicly available to the community. 
\section*{Acknowledgments}
This work utilized the Alpine high performance computing resource and the Blanca condo computing resource at the University of Colorado Boulder. Alpine is jointly funded by the University of Colorado Boulder, the University of Colorado Anschutz, Colorado State University, and the National Science Foundation (award 2201538). Blanca is jointly funded by computing users and the University of Colorado Boulder.

\bibliography{custom}

\appendix

\section{ILP Formulation}~\label{appendix:inference}
We formulate the ILP objective as follows:

\begin{equation}
\argmax_{\forall p\in P} \sum_{j}\sum_{k} w_{jk} \cdot p_{jk}
\end{equation}

\noindent Here, each $p \in P$ corresponds to a particular prompting strategy, and each $p_{jk}$ is a binary variable that represents whether a particular answer $k$ to an instance $j$ is true. Each $w_{jk}$ is the probability that the corresponding $p_{jk}$ is true. The weight itself is obtained by prompting an LLM using the prompting strategy $p$ and extracting a confidence value, using one of the strategies discussed in Section \ref{sec:scoring-strategies}. 

To support the use of different prompting strategies to score the same decisions, we introduce variables $d_{jk}$. These variables are not used in the objective function directly, but are used in defining constraints. Similarly to previous work \citep{zhang-etal-2016-introducing,pacheco-goldwasser-2021-modeling}, we define a set of standard constraints that can be used in different structured prediction tasks.

\paragraph{Multi-class constraints} For multiclass problems, only one variable can be activated among all possible variables for a multi-class decision on a particular instance decision.
\begin{equation}
\sum_k d_{jk}  = 1
\end{equation}

\paragraph{Decision constraints} It could be that, for a particular decision, we prompt an LLM in multiple different ways and/or with different contextual information. In order to obtain a single answer given multiple estimations, we need to constrain the assignment of the variables associated with that decision. We define a constraint to ensure that if a decision variable $d_{jk}$ is activated, at least one of its outcome variables must be activated.

\begin{equation}
 d_{jk}\leq \sum_{p}^{P} p_{jk}
\end{equation}

\noindent Where $P$ is the set of all prompting strategies. Conversely, the activation of any of the outcome variables associated with a given decision variable ensures the activation of that decision variable.

\begin{equation}
d_{jk} \geq \prod_{p}^{P} p_{jk}
\end{equation}

\paragraph{Hard constraints} We can define hard constraints to infuse domain knowledge. These constraints can be modeled in the form of ``if-then'' style rules. For example, let $d_{12,\text{coref}}$ 
be a variable that represents whether entities $e_1$ and $e_2$ are co-referent. We can ensure a transitivity constraint of the form \textit{if} $d_{12,\text{coref}}$ \textit{and} $d_{23,\text{coref}}$ \textit{then} $d_{13,\text{coref}}$ as:

\begin{equation}
d_{12,\text{coref}} + d_{23,\text{coref}} - 1 \leq d_{13,\text{coref}}
\end{equation}

\noindent Note that this generalizes for any horn clause $d_1 \wedge d_2 \wedge ... \wedge d_n \Rightarrow d_h$ as:

\begin{equation}
\sum^n_{i=1} d_i - (n - 1) \leq d_h
\end{equation}


\section{Morality Frames Prompt Details}\label{appendix:mf-prompt-details}
We split the Morality Frames structured prediction task into two subproblems. The prompt details for each of our prompting strategies when applied to the moral foundation classification subproblem can be found in Tables \ref{mf-white-box-prompt-details} and \ref{mf-black-box-prompt-details}. The prompt details for the moral role classification subproblem can be found in Tables \ref{mr-white-box-prompt-details} and \ref{mr-black-box-prompt-details}. For each subproblem we actually prompt in two different ways. In the first template we provide just the tweet itself as the [Tweet Context]. In the second template we provide the tweet, the political ideology of the tweet author, and the topic of the tweet as the [Tweet Context]. Additionally, for all strategies we provide the definitions of the associated moral foundations and roles.

\begin{table*}[t]
\vspace{0.2cm}
\centering
\resizebox{\textwidth}{!}{%
\begin{tabular}{|p{4cm}|p{12cm}|}
 \hline
\textbf{Prompting Strategy} & \textbf{Prompt Format}\\
\hline
True/False &
Consider the task of identifying the moral foundation present in a tweet from a U.S congress member. The five moral foundations and their corresponding definitions are given below:\newline

[Moral Frame Definitions]\newline

Given the moral foundations, their definitions, and the task of identifying the foundation present in a tweet, answer the following true/false question regarding whether a specific moral foundation is present in a tweet.\newline

[Tweet Context]\newline
Q. "The moral foundation expressed in the tweet is [label]." - true or false? A.\newline
\\
\hline
Multiple Choice &
Consider the task of identifying the moral foundation present in a tweet from a U.S congress member. The five moral foundations and their corresponding definitions are given below:\newline

[Moral Frame Definitions]\newline

Given the moral foundations, their definitions, and the task of identifying the foundation present in a tweet, answer the following multiple choice questions regarding whether a specific moral foundation is present in a tweet. Answer only with the letter corresponding to the correct answer.\newline

[Tweet Context]\newline
Q. What moral foundation is being expressed in the given tweet?\newline
Choices:\newline
(A) CARE/HARM\newline
(B) FAIRNESS/CHEATING\newline
(C) AUTHORITY/SUBVERSION\newline
(D) PURITY/DEGRADATION\newline
(E) LOYALTY/BETRAYAL\newline
\\
\hline
Generative Classification & 
Consider the task of generating a tweet made by a U.S congress member given a description of the moral foundation that is being expressed in the tweet. The five moral foundations and their corresponding definitions are given below:\newline

[Moral Frame Definitions]\newline

Given the moral foundations and their definitions, generate a tweet given a description of the tweet.\newline
Generate a tweet based on the following description:\newline
Generation description: [Generation Description]\newline
Tweet: [Tweet]\\
\hline
\end{tabular}}
\caption{Prompt templates for the moral foundation classification subproblem for each of the white-box strategies (Section \ref{sec:white-bot-methods}). [Generation Description] refers to one of the generation descriptions found in Table \ref{mf-generation-descriptions}.[label] is one of the five moral foundations (Care/Harm, Fairness/Cheating, Loyalty/Betrayal, Authority/Subversion, and Purity/Degradation).}
\label{mf-white-box-prompt-details}
\end{table*}

\begin{table*}[t]
\vspace{0.2cm}
\centering
\resizebox{\textwidth}{!}{%
\begin{tabular}{|p{4cm}|p{12cm}|}
 \hline
\textbf{Prompting Strategy} & \textbf{Prompt Format}\\
\hline
Generation Sampling & 
Consider the task of identifying the moral foundation present in a tweet from a U.S congress member. The five moral foundations and their corresponding definitions are given below:\newline

[Moral Frame Definitions]\newline

Given the moral foundations and their definitions, identify the foundation present in the given tweet. Only answer with the correct moral foundation and do not provide any justification or explanation.\newline
[Tweet Context] \newline
Q. What moral foundation is being expressed in the given tweet?\newline
\\
\hline
Verbalized Confidence & 
Consider the task of identifying the moral foundation present in a tweet. The five moral foundations and their corresponding definitions are given below:\newline

[Moral Frame Definitions]\newline

Given the moral foundations and their definitions and the task of identifying the foundation present in a tweet. Estimate the probability that the specified moral foundation is expressed in the tweet. Please answer with the following format:\newline
“Confidence: [the probability of answer [label] to be correct (0-100), not the one you think correct, please only include the numerical number in the range of 0-100]”\newline

Question: What is the moral foundation present in the following tweet: [Tweet Context]?\newline
 
Possible Answer: [label]\newline
 
Q: How likely is the above answer to be correct? Do not elaborate on your answer or provide any explantion, answer only with the confidence value in the following format:\newline
 
“Confidence: [the probability of answer [label] to be correct (0-100), not the one you think correct, please only include the numerical number in the range of 0-100]”\newline
\\
\hline
\end{tabular}}
\caption{Prompt templates for the moral foundation classification subproblem for each of the black-box strategies (Section \ref{sec:black-box-methods}). [label] is one of the five moral foundations (Care/Harm, Fairness/Cheating, Loyalty/Betrayal, Authority/Subversion, and Purity/Degradation).}
\label{mf-black-box-prompt-details}
\end{table*}

\begin{table*}[t]
\vspace{0.2cm}
\centering
\resizebox{\textwidth}{!}{%
\begin{tabular}{|p{4cm}|p{12cm}|}
 \hline
\textbf{Prompting Strategy} & \textbf{Prompt Format}\\
\hline
True/False &
Consider the task of identifying the moral role of an entity present in a tweet from a U.S congress member. Definitions for the five moral foundations and their associated roles are given below:\newline

[Moral Frame/Role Definitions]\newline

Given the possible moral roles, the definitions of their associated moral foundations, and the task of identifying the moral role of an entity in a tweet, answer the following true/false question regarding whether an entity is expressing a particular moral role in a tweet.\newline
[Tweet Context] \newline
Q. "The moral role of "[Entity]" expressed in the tweet is [label]." - true or false? A.\\
\hline
Multiple Choice &
Consider the task of identifying the moral role of an entity present in a tweet from a U.S congress member. Definitions for the five moral foundations and their associated roles are given below:\newline

[Moral Frame/Role Definitions]\newline

Given the possible moral roles, the definitions of their associated moral foundations, and the task of identifying the moral role of an entity in a tweet, answer the following multiple choice question regarding whether an entity is expressing a particular moral role in a tweet. Answer only with the letter corresponding to the correct answer.\newline

[Tweet Context]\newline
Q. What is the moral role of "[Entity]" expressed in the given tweet?\newline
(A) Target of care/harm\newline
(B) Entity causing harm\newline
(C) Entity providing care\newline
(D) Target of fairness/cheating\newline
(E) Entity ensuring fairness\newline
...\\
\hline
Generative Classification & 
Consider the task of generating a tweet made by a U.S congress member given a description of the moral role being expressed by an entity in the tweet. Definitions for the five moral foundations and their associated moral roles are given below:\newline

[Moral Frame/Role Definitions]\newline

Given the possible moral roles and the definitions of their associated moral foundations, generate a tweet about a given entity expressing a particular moral role.\newline
Generate a tweet based on the following description:\newline
Generation description: [Generation Description]\newline
Tweet: [Tweet]\\
\hline
\end{tabular}}
\caption{Prompt templates for the moral role identification subproblem for each of the white-box strategies (Section \ref{sec:white-bot-methods}). [Generation Description] refers to one of the generation descriptions found in Table \ref{mr-generation-descriptions}.[label] is one of the sixteen possible moral roles (Target of care/harm, Entity causing harm, Entity providing care, Target of fairness/cheating, Entity ensuring fairness, Entity doing cheating, Target of loyalty/betrayal, Entity being loyal, Entity doing betrayal, Justified authority, Justified authority over, Failing authority, Failing authority over, Target of purity/degradation, Entity preserving purity, Entity causing degradation).}
\label{mr-white-box-prompt-details}
\end{table*}

\begin{table*}[t]
\vspace{0.2cm}
\centering
\resizebox{\textwidth}{!}{%
\begin{tabular}{|p{4cm}|p{12cm}|}
 \hline
\textbf{Prompting Strategy} & \textbf{Prompt Format}\\
\hline
Generation Sampling & 
Consider the task of identifying the moral role of an entity present in a tweet from a U.S congress member. Definitions for the five moral foundations and their associated roles are given below:\newline

[Moral Frame/Role Definitions]\newline

Given the possible moral roles and the definitions of their associated moral foundations, identify the moral role of an entity in a tweet. Only answer with the correct moral role for the entity and do not provide any justification or explanation.\newline
[Tweet Context]\newline
Q. What is the moral role of "[Entity]" expressed in the given tweet?\\
\hline
Verbalized Confidence & 
Consider the task of identifying the moral role of an entity present in a tweet. Definitions for the five moral foundations and their associated roles are given below:\newline

[Moral Frame/Role Definitions]\newline

Given the possible moral roles, the definitions of their associated moral foundations, and the task of identifying the moral role of an entity in a tweet. Please answer with the following format: \newline
“Confidence: [the probability of answer [label] to be correct (0-100), not the one you think correct, please only include the numerical number in the range of 0-100]”\newline

Question: What is the moral role of the entity "[Entity]" expressed in the following tweet: [Tweet Context]?\newline
 
Possible Answer: [label]\newline
 
Q: How likely is the above answer to be correct? Do not elaborate on your answer or provide any explantion, answer only with the confidence value in the following format:\newline
 
“Confidence: [the probability of answer [label] to be correct (0-100), not the one you think correct, please only include the numerical number in the range of 0-100]”\newline
\\
\hline
\end{tabular}}
\caption{Prompt templates for the moral role identification subproblem for each of the back-box strategies (Section \ref{sec:black-box-methods}). [label] is one of the sixteen possible moral roles (Target of care/harm, Entity causing harm, Entity providing care, Target of fairness/cheating, Entity ensuring fairness, Entity doing cheating, Target of loyalty/betrayal, Entity being loyal, Entity doing betrayal, Justified authority, Justified authority over, Failing authority, Failing authority over, Target of purity/degradation, Entity preserving purity, Entity causing degradation).}
\label{mr-black-box-prompt-details}
\end{table*}

\begin{table*}[t]
\vspace{0.2cm}
\centering
\resizebox{\textwidth}{!}{%
\begin{tabular}{|p{16cm}|}
 \hline
\textbf{Generation Description}\\
\hline
"This tweet expresses the moral foundation [Moral Foundation] which is defined as: [Moral Foundation Definition]"\\
\hline
"This tweet reflects the moral foundation [Moral Foundation], which is defined as: [Moral Foundation Definition]"\\
\hline
"The tweet showcases the moral foundation [Moral Foundation], described as: [Moral Foundation Definition]"\\
\hline
"In this tweet, the moral foundation [Moral Foundation] is expressed, defined as: [Moral Foundation Definition]"\\
\hline
"This tweet highlights the moral foundation [Moral Foundation], which means: [Moral Foundation Definition]"\\
\hline
"The moral foundation [Moral Foundation] is conveyed in this tweet, defined as: [Moral Foundation Definition]"\\
\hline
"This tweet demonstrates the moral foundation [Moral Foundation], described as: [Moral Foundation Definition]"\\
\hline
"In this tweet, the author expresses the moral foundation [Moral Foundation], which is defined as: [Moral Foundation Definition]"\\
\hline
"This tweet communicates the moral foundation [Moral Foundation], described as: [Moral Foundation Definition]"\\
\hline
"This tweet conveys the moral foundation [Moral Foundation], defined as: [Moral Foundation Definition]"
\\
\hline
\end{tabular}}
\caption{Generation descriptions used for the generative classification strategy on the morality foundation classification subproblem. As with \citet{kumar2024genz}, we generated these variations using ChatGPT with the following prompt: Write 10 paraphrases of this sentence as a Python list. “This tweet expresses the moral foundation [Moral Foundation] which is defined as [Moral Foundation Definition].”}
\label{mf-generation-descriptions}
\end{table*}

\begin{table*}[t]
\vspace{0.2cm}
\centering
\resizebox{\textwidth}{!}{%
\begin{tabular}{|p{16cm}|}
 \hline
\textbf{Generation Description}\\
\hline
"In this tweet, the entity [Entity] displays the moral role [Moral Role], defined as: [Moral Role Definition]"\\
\hline
"This tweet shows the entity [Entity] exhibiting the moral role [Moral Role], which is defined as: [Moral Role Definition]"\\
\hline
"The entity [Entity] in this tweet demonstrates the moral role [Moral Role], described as: [Moral Role Definition]"\\
\hline
"In this tweet, [Entity] reflects the moral role [Moral Role], which is defined as: [Moral Role Definition]"\\
\hline
"[Entity] in this tweet exemplifies the moral role [Moral Role], defined as: [Moral Role Definition]"\\
\hline
"This tweet portrays the entity [Entity] as embodying the moral role [Moral Role], described as: [Moral Role Definition]"\\
\hline
"The entity [Entity] in this tweet illustrates the moral role [Moral Role], which is defined as: [Moral Role Definition]"\\
\hline
"[Entity] shows the moral role [Moral Role] in this tweet, defined as: [Moral Role Definition]"\\
\hline
"In this tweet, [Entity] reveals the moral role [Moral Role], defined as: [Moral Role Definition]"\\
\hline
"This tweet features [Entity] expressing the moral role [Moral Role], which is defined as: [Moral Role Definition]"
\\
\hline
\end{tabular}}
\caption{Generation descriptions used for the generative classification strategy on the morality role identification subproblem. As with \citet{kumar2024genz}, we generated these variations using ChatGPT with the following prompt: Write 10 paraphrases of this sentence as a Python list. “This entity [Entity] in this tweet exhibits the moral role [Moral Role] defined as [Moral Role Definition].}
\label{mr-generation-descriptions}
\end{table*}

\section{Morality Frames Hyperparameter Details}\label{appendix:mf-hyperparam}
For all strategies in the few shot case we use a topk of 5 and a temperature of 0.5. For the generation sampling and verbalized confidence strategies, we sample 10 generations per each instance of a subproblem. For the generative classification method we use 10 different variations of a generation description as found in Tables \ref{mf-generation-descriptions} and \ref{mr-generation-descriptions}.\\

For all fine-tuning strategies, we tune our hyperparamaters based on the average performance on the dev split over five fold cross validation. For the logistic regression models using the pre-trained LLM model scores as input, we use a learning rate of 0.01 and batch size of 32 for both the locally and globally calibrated models. For the fine-tuned LLM model scores, our regression models use a learning rate of 0.001 and batch size of 64 when locally calibrated and a learning rate of 0.01 and batch size of 16 when globally calibrated. For the supervised fine tuning of the LLM itself, we use LoRA \citep{hu2022lora} with a rank and alpha value of 512, a batch size of 2, gradient accumulation steps of 16 and a learning rate of 2E-05. For the structured calibration we continue tuning the same LoRA weights with a batch size of 4, gradient accumulation steps of 16, and a learning rate of 2E-06. 

\section{Extended Morality Frames Results}\label{appendix:extended-mf-results}

Tables \ref{mf-full-results-llama} and \ref{mf-full-results-mistral} show the full results for the llama and mistral models across all five prompting strategies (Section \ref{sec:scoring-strategies}) on the Morality Frames structured prediction task \citep{roy-etal-2021-identifying}.

\begin{table*}[t]
\vspace{0.2cm}
\centering
\begin{tabular}{llllllll}
 \hline
\textbf{Method} & \textbf{Shots} & \multicolumn{2}{c}{\textbf{Micro F1}}& \multicolumn{2}{c}{\textbf{Macro F1}}  & \multicolumn{2}{c}{\textbf{Constraint Violations}} \\
~ & ~ & \textbf{MF} & \textbf{Role} & \textbf{MF} & \textbf{Role} & \textbf{C1} & \textbf{C2}\\
 \hline
 \hline
 True/False & \multirow{2}{*}{0} & 0.369 & 0.347 & 0.401 & 0.247 & 1079 & 330 \\
 \hspace{3mm}+ constr & & \underline{0.429} & \textbf{\underline{0.362}} & \underline{0.437} & \textbf{\underline{0.293}} & 0 & 0 \\
 \hline
 Multiple Choice & \multirow{2}{*}{0} & 0.403 & 0.291 & 0.374 & 0.184 & 830 & 501 \\
 \hspace{3mm}+ constr & & \underline{0.445} & \underline{0.332} & \underline{0.412} & \underline{0.231} & 0 & 0 \\
 \hline
 Generation Sampling & \multirow{2}{*}{0} & 0.326 & \underline{0.363} & 0.317 & 0.239 & 981 & 172 \\
 \hspace{3mm}+ constr & & \underline{0.371} & 0.351 & \underline{0.331} & \underline{0.263} & 0 & 0 \\
 \hline
 Verbalized Confidence & \multirow{2}{*}{0} & 0.416 & 0.261 & 0.418 & 0.180 & 1275 & 312 \\
 \hspace{3mm}+ constr & & \underline{0.435} & \underline{0.322} & \underline{0.437} & \underline{0.274} & 0 & 0 \\
 \hline
 Generative Classification & \multirow{2}{*}{0} & 0.410 & 0.274 & 0.426 & 0.211 & 1072 & 338 \\
 \hspace{3mm}+ constr & & \textbf{\underline{0.498}} & \underline{0.295} & \textbf{\underline{0.483}} & \underline{0.245} & 0 & 0 \\
\hline
 \hline
 True/False & \multirow{2}{*}{2} & 0.448 & 0.386 & 0.422 & 0.313 & 1053 & 215 \\
 \hspace{3mm}+ constr & & \underline{0.486} & \textbf{\underline{0.391}} & \underline{0.467} & \textbf{\underline{0.346}} & 0 & 0 \\
 \hline
 Multiple Choice & \multirow{2}{*}{2} & 0.457 & 0.348 & 0.447 & 0.273 & 1133 & 234 \\
 \hspace{3mm}+ constr & & \textbf{\underline{0.530}} & \underline{0.386} & \textbf{\underline{0.512}} & \underline{0.321} & 0 & 0 \\
 \hline
 Generation Sampling & \multirow{2}{*}{2} & \underline{0.461} & 0.354 & \underline{0.492} & 0.271 & 1113 & 155 \\
 \hspace{3mm}+ constr & & 0.451 & \underline{0.357} & 0.475 & \underline{0.324} & 0 & 0 \\
 \hline
 \hline
  True/False & \multirow{2}{*}{5} & 0.498 & 0.444 & 0.466 & 0.378 & 1195 & 102 \\
 \hspace{3mm}+ constr & & \underline{0.517} & \textbf{\underline{0.452}} & \textbf{\underline{0.521}} & \textbf{\underline{0.408}} & 0 & 0 \\
 \hline
 Multiple Choice & \multirow{2}{*}{5} & 0.451 & 0.319 & 0.454 & 0.219 & 1254 & 219 \\
 \hspace{3mm}+ constr & & \textbf{\underline{0.528}} & \underline{0.361} & \underline{0.516} & \underline{0.277} & 0 & 0 \\
 \hline
 Generation Sampling & \multirow{2}{*}{5} & 0.389 & 0.387 & 0.419 & 0.276 & 1241 & 128 \\
 \hspace{3mm}+ constr & & \underline{0.456} & \underline{0.388} & \underline{0.447} & \underline{0.317} & 0 & 0 \\
 \hline
\end{tabular}
\caption{
Full Results for the Morality Frames task \citep{roy-etal-2021-identifying} using Llama-3.1-8B-Instruct for each of the five confidence elicitation methods (Section \ref{sec:scoring-strategies}), both with and without combinatorial inference.}
\label{mf-full-results-llama}
\end{table*}

\begin{table*}[t]
\vspace{0.2cm}
\centering
\begin{tabular}{llllllll}
 \hline
\textbf{Method} & \textbf{Shots} & \multicolumn{2}{c}{\textbf{Micro F1}}& \multicolumn{2}{c}{\textbf{Macro F1}}  & \multicolumn{2}{c}{\textbf{Constraint Violations}} \\
~ & ~ & \textbf{MF} & \textbf{Role} & \textbf{MF} & \textbf{Role} & \textbf{C1} & \textbf{C2}\\
 \hline
 \hline
 True/False & \multirow{2}{*}{0} & 0.352 & 0.350 & 0.274 & \textbf{\underline{0.213}} & 974 & 459 \\
 \hspace{3mm}+ constr & & \underline{0.457} & \underline{0.353} & \textbf{\underline{0.352}} & 0.206 & 0 & 0 \\
 \hline
 Multiple Choice & \multirow{2}{*}{0} & 0.407 & 0.290 & 0.207 & \underline{0.154} & 645 & 400 \\
 \hspace{3mm}+ constr & & \underline{0.415} & \textbf{\underline{0.386}} & \underline{0.251} & 0.145 & 0 & 0 \\
 \hline
 Generation Sampling & \multirow{2}{*}{0} & 0.347 & \underline{0.329} & 0.214 & \underline{0.195} & 1133 & 306 \\
 \hspace{3mm}+ constr & & \underline{0.382} & 0.262 & \underline{0.351} & 0.192 & 0 & 0 \\
 \hline
 Verbalized Confidence & \multirow{2}{*}{0} & 0.412 & 0.238 & 0.239 & \underline{0.136} & 617 & 547 \\
 \hspace{3mm}+ constr & & \textbf{\underline{0.430}} & \underline{0.267} & \underline{0.248} & 0.132 & 0 & 0 \\
 \hline
 Generative Classification & \multirow{2}{*}{0} & 0.374 & 0.189 & 0.333 & 0.140 & 1307 & 346 \\
 \hspace{3mm}+ constr & & \underline{0.415} & \underline{0.217} & \underline{0.365} & \underline{0.183} & 0 & 0 \\
\hline
 \hline
 True/False & \multirow{2}{*}{2} & 0.444 & 0.427 & 0.317 & 0.306 & 1332 & 197 \\
 \hspace{3mm}+ constr & & \textbf{\underline{0.528}} & \textbf{\underline{0.436}} & \textbf{\underline{0.495}} & \textbf{\underline{0.358}} & 0 & 0 \\
 \hline
 Multiple Choice & \multirow{2}{*}{2} & 0.453 & 0.317 & 0.433 & 0.201 & 1159 & 213 \\
 \hspace{3mm}+ constr & & \underline{0.491} & \underline{0.367} & \underline{0.452} & \underline{0.252} & 0 & 0 \\
 \hline
 Generation Sampling & \multirow{2}{*}{2} & 0.401 & \underline{0.354} & 0.388 & 0.233 & 1348 & 158 \\
 \hspace{3mm}+ constr & & \underline{0.440} & 0.333 & \underline{0.417} & \underline{0.252} & 0 & 0 \\
 \hline
 \hline
  True/False & \multirow{2}{*}{5} & 0.500 & \textbf{\underline{0.464}} & 0.483 & 0.354 & 1214 & 151 \\
 \hspace{3mm}+ constr & & \textbf{\underline{0.536}} & 0.426 & \textbf{\underline{0.513}} & \textbf{\underline{0.377}} & 0 & 0 \\
 \hline
 Multiple Choice & \multirow{2}{*}{5} & 0.458 & 0.332 & \underline{0.433} & 0.224 & 1109 & 228 \\
 \hspace{3mm}+ constr & & \underline{0.498} & \underline{0.384} & 0.426 & \underline{0.246} & 0 & 0 \\
 \hline
 Generation Sampling & \multirow{2}{*}{5} & 0.357 & \underline{0.368} & \underline{0.386} & 0.214 & 1464 & 180 \\
 \hspace{3mm}+ constr & & \underline{0.412} & 0.346 & 0.376 & \underline{0.230} & 0 & 0 \\
 \hline
\end{tabular}
\caption{
Full Results for the Morality Frames task \citep{roy-etal-2021-identifying} using Mistral-7B-Instruct-v0.2 for each of the five confidence elicitation methods (Section \ref{sec:scoring-strategies}), both with and without combinatorial inference.}
\label{mf-full-results-mistral}
\end{table*}

\section{Error Analysis}\label{appendix:error-analysis}

\begin{figure*}\includegraphics[width=\textwidth]{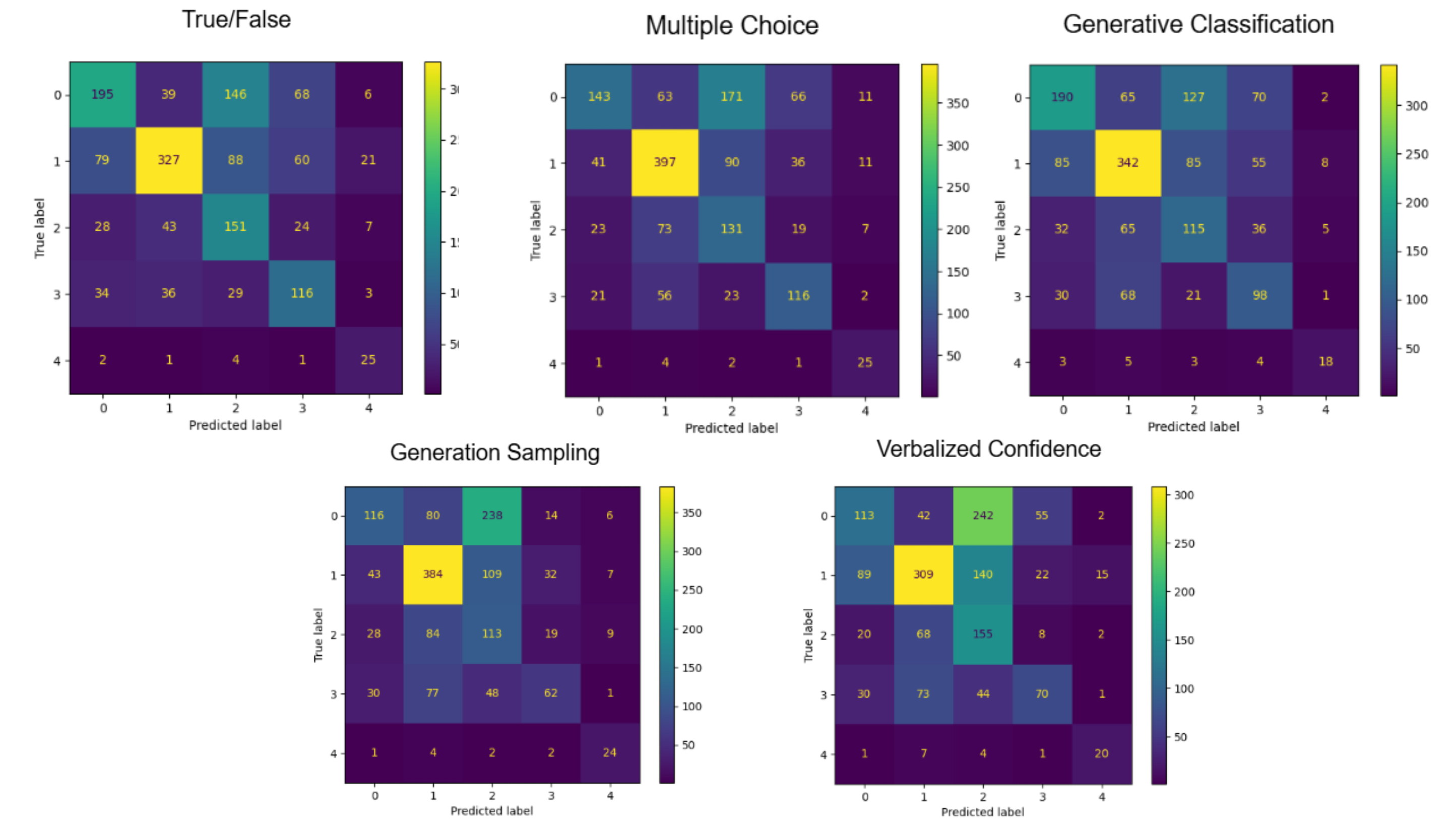} \caption{Confusion matrices for the morality frames task} \label{fig:confusionmatricies} \end{figure*}

Confusion matrices for the moral framing task can be seen in Figure~\ref{fig:confusionmatricies}.

\section{Coref Prompt Details}\label{appendix:coref-prompt-details}
Tables \ref{coref-prompt-details-white-box} and \ref{coref-prompt-details-black-box} display the prompt templates used in the coreference tasks for the white box (Section \ref{sec:white-bot-methods}) and black box (Section \ref{sec:black-box-methods}) prompting strategies respectively. Each strategy relies on the sentences that contain each entity mention being compared ([sent1], [sent2]) and the entity mentions themselves ([entity1], [entity2]). Additionally, the generation description strategy uses one of the variations in Table \ref{coref-prompt-details-generation-descriptions} as generation instructions.
\begin{table*}[t]
\vspace{0.2cm}
\centering
\resizebox{\textwidth}{!}{%
\begin{tabular}{|p{4cm}|p{12cm}|}
 \hline
\textbf{Prompting Strategy} & \textbf{Prompt Format}\\
\hline
True/False & Consider the task of coreference resolution, where the goal is to identify whether or not two different entity mentions refer to the same underlying entity. Given two entity mentions and their representative sentences, answer the following true/false question regarding whether the two entity mentions refer to the same entity.\newline
Entity 1: [entity1]\newline
Sentence 1: [sent1]\newline
Entity 2: [entity2]\newline
Sentence 2: [sent2]\newline

Q. "The entity "[entity1]" mentioned in Sentence 1 and the entity "[entity2]" mentioned in Sentence 2 are [label] entities." - true or false? A.\\
\hline
Multiple Choice & 
Consider the task of coreference resolution, where the goal is to identify whether or not two different entity mentions refer to the same underlying entity. Given two entity mentions and their representative sentences, answer the following multiple choice question regarding whether or not the two entity mentions are coreferent or not. Answer only with the letter corresponding to the correct answer.\newline
Entity 1: [entity1]\newline
Sentence 1: [sent1]\newline
Entity 2: [entity2]\newline
Sentence 2: [sent2]\newline

Q. What is the relationship between the entity "[entity1]" mentioned in Sentence 1 and "[entity2]" mentioned in Sentence 2? \newline
(A) Coreferent\newline
(B) Distinct\\
\hline
Generative Classification & 
Consider the task of coreference resolution. Given two entities mentions that are either coreferent or distinct, generate two sentences each containing one of the entity mentions.\newline
Generate two sentences based on the following description:\newline

Generation Description: [Generation Description]\newline
Sentence 1: [sent1]\newline
Sentence 2: [sent2]\newline\\
\hline
\end{tabular}}
\caption{Coreference prompt templates for each of the white-box prompting strategies (Section \ref{sec:white-bot-methods}). The [Generation Description] variable refers to one of the generation descriptions in Table \ref{coref-prompt-details-generation-descriptions}. [label] is either `coreferent' or `distinct'. }
\label{coref-prompt-details-white-box}
\end{table*}

\begin{table*}[t]
\vspace{0.2cm}
\centering
\resizebox{\textwidth}{!}{%
\begin{tabular}{|p{4cm}|p{12cm}|}
 \hline
\textbf{Prompting Strategy} & \textbf{Prompt Format}\\
\hline
Generation Sampling & 
Consider the task of coreference resolution, where the goal is to identify whether or not two different entity mentions refer to the same underlying entity. Given two entity mentions and their representative sentences, identify whether the entity mentions are coreferent or distinct. Answer only with "coreferent" or "distinct" and do not provide any justification or explanation.\newline
Entity 1: [entity1]\newline
Sentence 1: [sent1]\newline
Entity 2: [entity2]\newline
Sentence 2: [sent2]\newline

Q. What is the relationship between the entity "[entity1]" mentioned in Sentence 1 and "[entity2]" mentioned in Sentence 2? Answer only with "coreferent" or "distinct" and do not provide any justification or explanation.\\
\hline
Verbalized Confidence & 
Consider the task of coreference resolution, where the goal is to identify whether or not two different entity mentions refer to the same underlying entity. Given two entity mentions and their representative sentences, identify whether the entity mentions are coreferent or distinct. Please answer with the following format:
“Confidence: [the probability that the two entity mentions are [label] (0-100), please only include the numerical number in the range of 0-100]”\newline
Entity 1: [entity1]\newline
Sentence 1: [sent1]\newline
Entity 2: [entity2]\newline
Sentence 2: [sent2]\newline

Q: How likely is it that the two entitiy mentions are [label]. Do not elaborate on your answer or provide any explantion, answer only with the confidence value in the following format: \newline

“Confidence: [the probability that the two entity mentions are [label] (0-100), please only include the numerical number in the range of 0-100]”\\
\hline
\end{tabular}}
\caption{Coreference prompt templates for each of the black-box prompting strategies (Section \ref{sec:black-box-methods}). [label] is either `coreferent' or `distinct'. }
\label{coref-prompt-details-black-box}
\end{table*}

\begin{table*}[t]
\vspace{0.2cm}
\centering
\resizebox{\textwidth}{!}{%
\begin{tabular}{|p{16cm}|}
 \hline
\textbf{Generation Description}\\
\hline
"The entity mention '[entity1]' in the first sentence and the entity mention of '[entity2]' in the second sentence are [label]"\\
\hline
"The mention of '[entity1]' in sentence one and '[entity2]' in sentence two are considered [label]."\\
\hline
"'[entity1]' from the first sentence and '[entity2]' from the second sentence are labeled as [label]."\\
\hline
"In the first sentence, '[entity1]' is mentioned, and in the second sentence, '[entity2]' is mentioned — these are [label]."\\
\hline
"The entities '[entity1]' and '[entity2]', from the first and second sentences respectively, are [label]."\\
\hline
"We determine that '[entity1]' in sentence one and '[entity2]' in sentence two are [label]."\\
\hline
"There is a mention of '[entity1]' in the first sentence and of '[entity2]' in the second; they are [label]."\\
\hline
"According to the sentence context, '[entity1]' and '[entity2]' are identified as [label]."\\
\hline
"It is determined that the entity '[entity1]' in the first sentence and '[entity2]' in the second sentence are [label]."\\
\hline
"In the first sentence, the mention of '[entity1]', and in the second, the mention of '[entity2]', are considered [label]."\\
\hline
\end{tabular}}
\caption{Generation descriptions used for the generative classification strategy (Section \ref{sec:generative-classification}). As with \citet{kumar2024genz}, we generated these variations using ChatGPT with the following prompt: Write 10 paraphrases of this sentence as a Python list. “The entity mention [entity1] in the first sentence and the entity mention of [entity2] in the second sentence are [label]."  }
\label{coref-prompt-details-generation-descriptions}
\end{table*}

\section{Extended Coref Results}\label{appendix:extended-coref-results}

Tables \ref{full-genia-results-llama} and \ref{full-genia-results-mistral} show the full results for the llama and mistral models across all five prompting strategies (Section \ref{sec:scoring-strategies}) on the GENIA coreference dataset \citep{su2008coreference}. Tables \ref{full-ontonotes-results-llama} and \ref{full-ontonotes-results-mistral} show the full results on the OntoNotes coreference dataset \citep{pradhan2012conll}.

\begin{table}[t]
\centering
\resizebox{\columnwidth}{!}{%
\begin{tabular}{llll}
\hline
\textbf{Method} & \textbf{Shots} & \textbf{F1} & \textbf{Viol.} \\
\hline
\hline
 True/False & \multirow{2}{*}{0} & 0.695 & 106502 \\
 \hspace{3mm}+ constr & & \textbf{\underline{0.758}} & 0 \\
 \hline
 Multiple Choice & \multirow{2}{*}{0} & \underline{0.427} & 345664 \\
 \hspace{3mm}+ constr & & 0.420 & 0 \\
 \hline
 Generation Sampling & \multirow{2}{*}{0} & 0.566 & 249450 \\
 \hspace{3mm}+ constr & & \underline{0.598} & 0 \\
 \hline
 Verbalized Confidence & \multirow{2}{*}{0} & \underline{0.321} & 338178 \\
 \hspace{3mm}+ constr & & 0.269 & 0 \\
 \hline
 Generative Classification & \multirow{2}{*}{0} & \underline{0.357} & 364712 \\
 \hspace{3mm}+ constr & & 0.305 & 0 \\
 \hline
 \hline
  True/False & \multirow{2}{*}{5} & 0.731 & 89982 \\
 \hspace{3mm}+ constr & & \textbf{\underline{0.797}} & 0 \\
 \hline
 Multiple Choice & \multirow{2}{*}{5} & 0.665 & 131086 \\
 \hspace{3mm}+ constr & & \underline{0.719} & 0 \\
 \hline
 Generation Sampling & \multirow{2}{*}{5} & 0.750 & 55980 \\
 \hspace{3mm}+ constr & & \underline{0.779} & 0 \\
 \hline
 \hline
\end{tabular}}
\caption{
Full Co-reference results for the GENIA dataset \citep{su2008coreference} using Llama-3.1-8B-Instruct across all five prompting strategies (Section \ref{sec:scoring-strategies}).
}\label{full-genia-results-llama}
\end{table}

\begin{table}[t]
\centering
\resizebox{\columnwidth}{!}{%
\begin{tabular}{llll}
\hline
\textbf{Method} & \textbf{Shots} & \textbf{F1} & \textbf{Viol.} \\
\hline
\hline
 True/False & \multirow{2}{*}{0} & 0.772 & 42054 \\
 \hspace{3mm}+ constr & & \textbf{\underline{0.795}} & 0 \\
 \hline
 Multiple Choice & \multirow{2}{*}{0} & \underline{0.600} & 163160 \\
 \hspace{3mm}+ constr & & 0.592 & 0 \\
 \hline
 Generation Sampling & \multirow{2}{*}{0} & 0.658 & 128512 \\
 \hspace{3mm}+ constr & & \underline{0.686} & 0 \\
 \hline
 Verbalized Confidence & \multirow{2}{*}{0} & \underline{0.599} & 123272 \\
 \hspace{3mm}+ constr & & 0.577 & 0 \\
 \hline
 Generative Classification & \multirow{2}{*}{0} & \underline{0.219} & 259610 \\
 \hspace{3mm}+ constr & & 0.150 & 0 \\
 \hline
 \hline
  True/False & \multirow{2}{*}{5} & 0.799 & 29808 \\
 \hspace{3mm}+ constr & & \textbf{\underline{0.823}} & 0 \\
 \hline
 Multiple Choice & \multirow{2}{*}{5} & 0.721 & 72303 \\
 \hspace{3mm}+ constr & & \underline{0.759} & 0 \\
 \hline
 Generation Sampling & \multirow{2}{*}{5} & 0.747 & 48570 \\
 \hspace{3mm}+ constr & & \underline{0.781} & 0 \\
 \hline
 \hline
\end{tabular}}
\caption{
Full Co-reference results for the GENIA dataset \citep{su2008coreference} using Mistral-7B-Instruct-v0.2 across all five prompting strategies (Section \ref{sec:scoring-strategies}).
}\label{full-genia-results-mistral}
\end{table}

\begin{table}[t]
\centering
\resizebox{\columnwidth}{!}{%
\begin{tabular}{llll}
\hline
\textbf{Method} & \textbf{Shots} & \textbf{F1} & \textbf{Viol.} \\
\hline
\hline
 True/False & \multirow{2}{*}{0} & 0.763 & 21264 \\
 \hspace{3mm}+ constr & & \textbf{\underline{0.774}} & 0 \\
 \hline
 Multiple Choice & \multirow{2}{*}{0} & \underline{0.476} & 53008 \\
 \hspace{3mm}+ constr & & 0.455 & 0 \\
 \hline
 Generation Sampling & \multirow{2}{*}{0} & \underline{0.661} & 51524 \\
 \hspace{3mm}+ constr & & 0.657 & 0 \\
 \hline
 Verbalized Confidence & \multirow{2}{*}{0} & \underline{0.423} & 44480 \\
 \hspace{3mm}+ constr & & 0.291 & 0 \\
 \hline
 Generative Classification & \multirow{2}{*}{0} & \underline{0.315} & 27062 \\
 \hspace{3mm}+ constr & & 0.272 & 0 \\
 \hline
 \hline
  True/False & \multirow{2}{*}{5} & 0.753 & 36884 \\
 \hspace{3mm}+ constr & & \underline{0.777} & 0 \\
 \hline
 Multiple Choice & \multirow{2}{*}{5} & 0.801 & 24928 \\
 \hspace{3mm}+ constr & & \underline{0.830} & 0 \\
 \hline
 Generation Sampling & \multirow{2}{*}{5} & 0.834 & 15086 \\
 \hspace{3mm}+ constr & & \textbf{\underline{0.842}} & 0 \\
 \hline
 \hline
\end{tabular}}
\caption{
Full Co-reference results for the OntoNotes dataset \citep{pradhan2012conll} using Llama-3.1-8B-Instruct across all five prompting strategies (Section \ref{sec:scoring-strategies}).
}\label{full-ontonotes-results-llama}
\end{table}

\begin{table}[t]
\centering
\resizebox{\columnwidth}{!}{%
\begin{tabular}{llll}
\hline
\textbf{Method} & \textbf{Shots} & \textbf{F1} & \textbf{Viol.} \\
\hline
\hline
 True/False & \multirow{2}{*}{0} & \textbf{\underline{0.747}} & 9540 \\
 \hspace{3mm}+ constr & & 0.729 & 0 \\
 \hline
 Multiple Choice & \multirow{2}{*}{0} & \underline{0.648} & 42444 \\
 \hspace{3mm}+ constr & & 0.627 & 0 \\
 \hline
 Generation Sampling & \multirow{2}{*}{0} & \underline{0.709} & 32312 \\
 \hspace{3mm}+ constr & & 0.702 & 0 \\
 \hline
 Verbalized Confidence & \multirow{2}{*}{0} & 0.506 & 25872 \\
 \hspace{3mm}+ constr & & \underline{0.512} & 0 \\
 \hline
 Generative Classification & \multirow{2}{*}{0} & \underline{0.397} & 47794 \\
 \hspace{3mm}+ constr & & 0.371 & 0 \\
 \hline
 \hline
  True/False & \multirow{2}{*}{5} & 0.815 & 15212 \\
 \hspace{3mm}+ constr & & \textbf{\underline{0.820}} & 0 \\
 \hline
 Multiple Choice & \multirow{2}{*}{5} & 0.741 & 32712 \\
 \hspace{3mm}+ constr & & \underline{0.748} & 0 \\
 \hline
 Generation Sampling & \multirow{2}{*}{5} & 0.792 & 21858 \\
 \hspace{3mm}+ constr & & \underline{0.799} & 0 \\
 \hline
 \hline
\end{tabular}}
\caption{
Full Co-reference results for the OntoNotes dataset \citep{pradhan2012conll} using Mistral-7B-Instruct-v0.2 across all five prompting strategies (Section \ref{sec:scoring-strategies}).
}\label{full-ontonotes-results-mistral}
\end{table}

\section{Coreference Hyperparameter Details}\label{appendix:coref-hyperparam}
For all strategies in the few shot case for both the GENIA and OntoNotes coreference datasets we use a topk of 5 and a temperature of 0.5. For the generation sampling and verbalized confidence strategies, we sample 10 generations per each instance of a subproblem. For the generative classification method, we use 10 different variations of a generation description as found in Table \ref{coref-prompt-details-generation-descriptions}.\\

For all fine-tuning strategies on the GENIA dataset we use the dev set for hyperparamater selection and report results on the test set. We use the same train/test split as \citet{mehta-etal-2024-promptly}. For the logistic regression models used on the pre-trained LLM scores, we use a learning rate of 0.001 and batch size of 32 for the locally calibrated models and a learning rate of 0.01 and batching done at the document level for the globally calibrated models. For the regression models trained on the fine-tuned scores, we use a learning rate of 0.01 and document batching for both locally and globally calibrated models. As with the Morality Frames task, we use LoRA for LLM fine-tuning with a rank and alpha of 512. For the supervised fine tuned model, we use a batch size of 2, gradient accumulation steps of 16, and a learning rate of 2E-6. For global fine tuning, we continue tuning the same LoRA weights with a learning rate of 1E-6, a batch size of 8, and gradient accumulation steps of 32.

\section{Hardware Details}\label{app:hardware}

We use a 40GB partition of an H100 GPU for all of the 7-8B pretrained model experiments and use a full 80GB H100 for training and inference experiments with the larger models. Regardless of GPU size, all experiments utilize 8 CPUs and 32GB of CPU RAM. 
\end{document}